\documentclass[11pt,a4paper]{article}
\usepackage{authblk}
\usepackage[hyperref]{acl2020}
\usepackage{times}
\usepackage{latexsym}

\usepackage{graphicx,calc,amsmath,upgreek}
\usepackage{microtype}
\usepackage{multirow}
\usepackage{booktabs}
\usepackage{amsmath}

\DeclareMathOperator*{\argmax}{arg\,max}

\aclfinalcopy 
\def\aclpaperid{2659} 
\title{Worse WER, but Better BLEU? Leveraging Word Embedding as Intermediate in Multitask End-to-End Speech Translation}

\author[1]{Shun-Po Chuang}
\author[2]{Tzu-Wei Sung}
\author[1]{Alexander H. Liu}
\author[1]{Hung-yi Lee}
\affil[1]{National Taiwan University, Taiwan}
\affil[2]{University of California San Diego, USA}
\affil[ ]{\small\texttt{\{f04942141, b03902042, r07922013, hungyilee\}@ntu.edu.tw}}

\date{}

\begin{document}
\maketitle
\begin{abstract}
Speech translation (ST) aims to learn transformations from speech in the source language to the text in the target language. Previous works show that multitask learning improves the ST performance, in which the recognition decoder generates the text of the source language, and the translation decoder obtains the final translations based on the output of the recognition decoder. Because whether the output of the recognition decoder has the correct semantics is more critical than its accuracy, we propose to improve the multitask ST model by utilizing word embedding as the intermediate.

\end{abstract}

\section{Introduction}
Speech translation (ST) increasingly receives attention from the machine translation (MT) community recently.
To learn the transformation between speech in the source language and the text in the target language, conventional models pipeline automatic speech recognition (ASR) and text-to-text MT model~\cite{berard2016listen}.
However, such pipeline systems suffer from error propagation.

Previous works show that deep end-to-end models can outperform conventional pipeline systems with sufficient training data~\cite{weiss2017sequence,inaguma2019multilingual,sperber2019attention}.
Nevertheless, well-annotated bilingual data is expensive and hard to collect~\cite{bansal2018low,bansal2018pre,duong2016attentional}.
Multitask learning plays an essential role in leveraging a large amount of monolingual data to improve representation in ST.
Multitask ST models have two jointly learned decoding parts, namely the recognition and translation part.
The recognition part firstly decodes the speech of source language into the text of source language, and then based on the output of the recognition part, the translation part generates the text in the target language.
Variant multitask models have been explored~\cite{anastasopoulos-chiang-2018-tied}, which shows the improvement in low-resource scenario.

Although applying the text of source language as the intermediate information in multitask end-to-end ST empirically yielded improvement, we argue whether this is the optimal solution. 
Even though the recognition part does not correctly transcribe the input speech into text, the final translation result would be correct if the output of the recognition part preserves sufficient semantic information for translation.
Therefore, we explore to leverage word embedding as the intermediate level instead of text.

In this paper, we apply pre-trained word embedding as the intermediate level in the multitask ST model.
We propose to constrain the hidden states of the decoder of the recognition part to be close to the pre-trained word embedding.
Prior works on word embedding regression show improved results on MT~\cite{jauregi-unanue-etal-2019-rewe,kumar2018mises}.
Experimental results show that the proposed approach obtains improvement to the ST model.
Further analysis also shows that constrained hidden states are approximately isospectral to word embedding space, indicating that the decoder achieves speech-to-semantic mappings.

\section{Multitask End-to-End ST model}
\label{sec:memodel}
\begin{figure}[!t]
    \centering
    \resizebox{\linewidth}{!}
    {
        \includegraphics[width=1\textwidth]{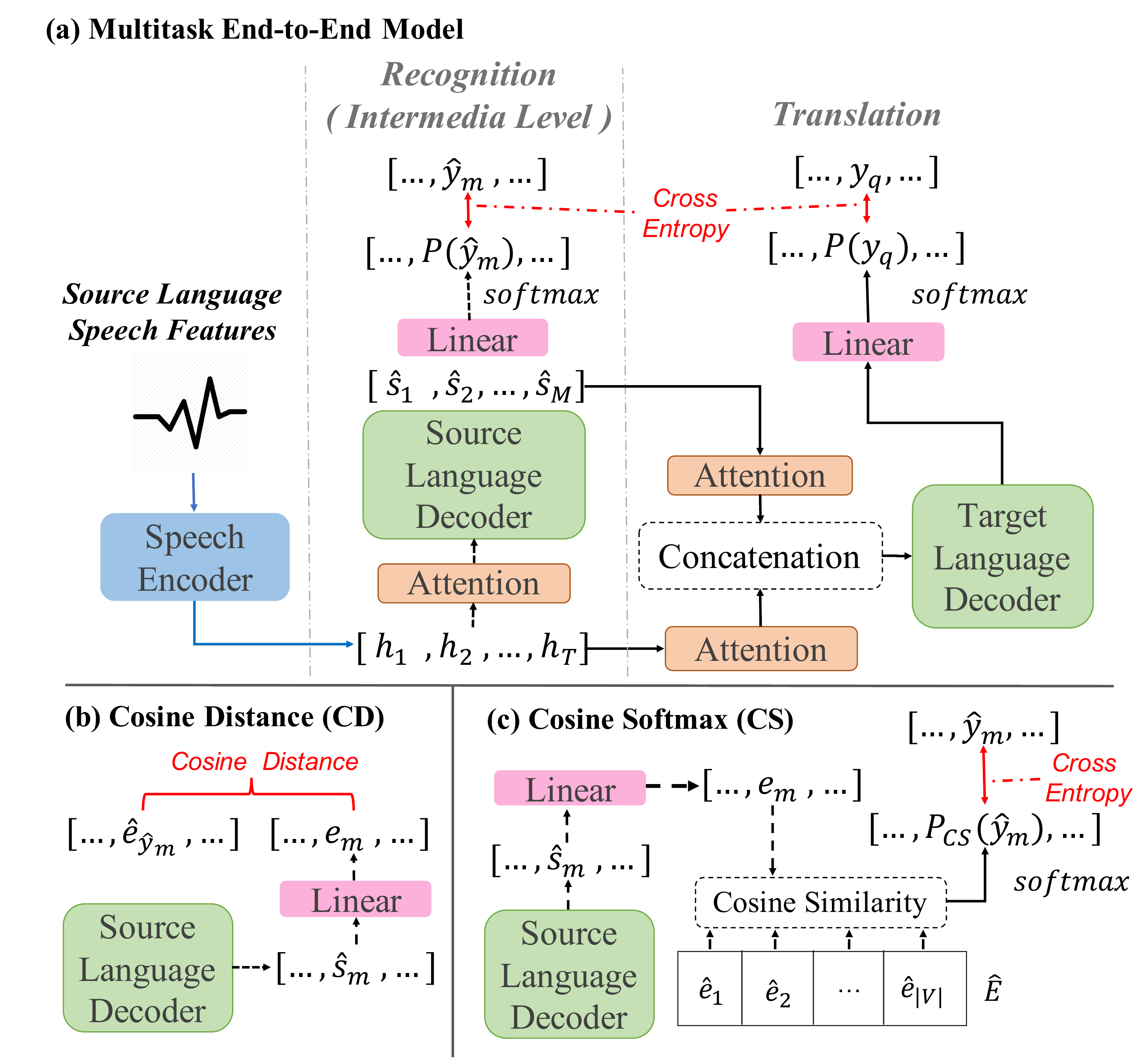}
    }
    \caption{(a) Multitask ST model. Dotted arrows indicate steps in the recognition part. Solid arrows indicate steps in the translation part. (b) Directly learn word embedding via cosine distance. (c) Learn word embedding via cosine softmax function. Both (b)(c) are the recognition part in (a).
    }
    \label{fig:triangle_model}
\end{figure}

Our method is based on the multitask learning for ST~\cite{anastasopoulos-chiang-2018-tied}, including speech recognition in the source language and translation in the target language, as shown in Fig.~\ref{fig:triangle_model}(a).
The input audio feature sequence is first encoded into the encoder hidden state sequence
${h = h_{1}, h_{2}, \dots, h_{T}}$
with length $T$ by the pyramid encoder~\cite{chan2015listen}.
To present speech recognition in the source language, the attention mechanism and a decoder is employed to produce source decoder sequence
${\hat{s} = \hat{s}_{1}, \hat{s}_{2}, \dots, \hat{s}_{M}}$, where $M$ is the number of decoding steps in the source language.
For each decoding step $m$, the probability $P(\hat{y}_m)$ of predicting the token $\hat{y}_m$ in the source language vocabulary can be computed based on the corresponding decoder state $\hat{s}_m$.

To perform speech translation in the target language, 
both the source language decoder state sequence $\hat{s}$ and the encoder state sequence $h$ will be attended and treated as the target language decoder's input.
The hidden state of target language decoder can then be used to derived the probability $P(y_q)$ of predicting token $y_q$ in the target language vocabulary for every decoding step $q$.

Given the ground truth sequence in the source language ${\hat{y} =\hat{y}_1, \hat{y}_2, \dots, \hat{y}_M}$ and the target language ${y=y_1, y_2, \dots, y_Q}$ with length $Q$,
multitask ST can be trained with maximizing log likelihood in \textit{both} domains.
Formally, the objective function of multitask ST can be written as:
\begin{equation} \label{eq:multitasking_loss}
    \begin{split}
    \mathcal{L}_\text{ST} =& \frac{\alpha}{M} \mathcal{L}_\text{src} + \frac{\beta}{Q}\mathcal{L}_\text{tgt}\\
    =& \frac{\alpha}{M} \sum_{m} -\log P(\hat{y}_m) + \frac{\beta}{Q} \sum_{q} -\log P(y_q),\\
    \end{split}
\end{equation}

\section{Proposed Methods} 
\label{proposedmethods}

We propose two ways to help the multitask end-to-end ST model capture the semantic relation between word tokens by leveraging the source language word embedding as intermediate level.
$\hat{E} = \{\hat{e}_{1}, \hat{e}_{2}, ... \hat{e}_{|V|}\}$, where $V$ is the vocabulary set and $\hat{e}_v \in \mathcal{R}^D$ is the embedding vector with dimension $D$ for any word $v \in V$, in the recognition task.
We choose the source language decoder state (embedding) $\hat{s}$ to reinforce since it is later used in the translation task.
To be more specific, we argue that the embedding generated by the source language decoder should be more \textit{semantically} correct in order to benefit the translation task.
Given the pre-trained source language word embedding $\hat{E}$, we proposed to constrain the source decoder state $\hat{s}_m$ at step $m$ to be close to its corresponding word embedding $\hat{e}_{\hat{y}_m}$ with the two approaches detailed in the following sections.

\subsection{Directly Learn Word Embedding}
\label{sec:cosine}
Since semantic-related words would be close in terms of cosine distance~\cite{mikolov2018advances}, a simple idea is to 
minimize the cosine distance (CD) between the source language decoder hidden state 
${\hat{s}_m}$ and the corresponding word embedding $\hat{e}_{\hat{y}_m}$ for every decode step $m$,
\begin{equation} \label{eq:cosine_distance}
\begin{split}
     \mathcal{L}_{\text{CD}} =& \sum_{m} 1 - \cos(f_\theta(\hat{s}_m),\hat{e}_{\hat{y}_m}) \\
    =& \sum_{m} 1 - \frac{f_\theta(\hat{s}_m) \cdot \hat{e}_{\hat{y}_m}}{\|f_\theta(\hat{s}_m)\|\| \hat{e}_{\hat{y}_m}\|},
\end{split}
\end{equation}
where $f_\theta(\cdot)$ is a learnable linear projection to match the dimensionality of word embedding and decoder state.
With this design, the network architecture of the target language decoder would not be limited by the dimension of word embedding. 
Fig.~\ref{fig:triangle_model}(b) illustrates this approach.
By replacing $\mathcal{L}_{\text{src}}$ in Eq.~(\ref{eq:multitasking_loss}) with $\mathcal{L}_{\text{CD}}$, semantic learning from word embedding for source language recognition can be achieved.

\subsection{Learn Word Embedding via Probability}
\label{sec:cosine_softmax}

Ideally, using word embedding as the learning target via minimizing CD can effectively train the decoder to model the semantic relation existing in the embedding space. However, such an approach suffers from the hubness problem~\cite{faruqui2016problems} of word embedding in practice (as we later discuss in Sec.~\ref{subsubsec:discussion}).

To address this problem, we introduce cosine softmax (CS) function~\cite{liu2017learning,liu2017rethinking} to learn speech-to-semantic embedding mappings.
Given the decoder hidden state $\hat{s}_m$ and the word embedding $\hat{E}$, the probability of the target word $\hat{y}_m$ is defined as
\begin{equation}
\label{eq:cos_softmax}
    P_\text{CS}(\hat{y}_{m}) = \frac{\exp(\cos(f_\theta(\hat{s}_m), \hat{e}_{\hat{y}_m}) / \uptau )}{\sum_{\hat{e}_v \in \hat{E}} \exp( \cos(f_\theta(\hat{s}_m), \hat{e}_v) / \uptau)},
\end{equation}
where $\cos(\cdot)$ and $f_\theta(\cdot)$ are from Eq.~(\ref{eq:cosine_distance}), and $\uptau$ is the temperature of softmax function.
Note that since the temperature $\uptau$ re-scales cosine similarity, the hubness problem can be mitigated by selecting a proper value for $\uptau$.
Fig.~\ref{fig:triangle_model}(c) illustrates the approach.
With the probability derived from cosine softmax in Eq.~(\ref{eq:cos_softmax}), the objective function for source language decoder can be written as
\begin{equation} \label{eq:cosine_softmax}
     \mathcal{L}_{\text{CS}}  = \sum_{m} -\log P_{\text{CS}}(\hat{y}_m).
\end{equation}
By replacing $\mathcal{L}_{\text{src}}$ in Eq.~(\ref{eq:multitasking_loss}) with $\mathcal{L}_{\text{CS}}$, the decoder hidden state sequence $\hat{s}$ is forced to contain semantic information provided by the word embedding.

\section{Experiments}
\label{results}
\subsection{Experimental Setup}
We used Fisher Spanish corpus~\cite{fishercorpus} to perform Spanish speech to English text translation. 
And we followed previous works~\cite{inaguma2019multilingual} for pre-processing steps, and 40/160 hours of \emph{train} set, standard \emph{dev}-\emph{test} are used for the experiments.
Byte-pair-encoding (BPE)~\cite{kudo2018sentencepiece} was applied to the target transcriptions to form 10K subwords as the target of the translation part.
Spanish word embeddings were obtained from FastText pre-trained on Wikipedia~\cite{bojanowski2016enriching}, and 8000 Spanish words were used in the recognition part.

The encoder is a 3-layer 512-dimensional bidirectional LSTM with additional convolution layers, yielding 8$\times$ down-sampling in time.
The decoders are 1024-dimensional LSTM, and we used one layer in the recognition part and two layers in the translation part.
The models were optimized using Adadelta with $10^{-6}$ as the weight decay rate.
Scheduled sampling with probability 0.8 was applied to the decoder in the translation part.
Experiments ran 1.5M steps, and models were selected by the highest BLEU on four transcriptions per speech in \emph{dev} set.

\subsection{Speech Translation Evaluation}
\label{subsec:evaluations}

\noindent \textbf{Baseline}: We firstly built the single-task end-to-end model (\textbf{SE}) to set a baseline for multitask learning, which resulted in 34.5/34.51 BLEU on \emph{dev} and \emph{test} set respectively, which showed comparable results to ~\citet{salesky2019exploring}.
Multitask end-to-end model (\textbf{ME}) mentioned in Sec.~\ref{sec:memodel} is another baseline.
By applying multitask learning in addition, we could see that \textbf{ME} outperforms \textbf{SE} in all conditions.

\begin{table}[t]
\centering
\begin{tabular}{c|cccc}
\toprule
\multirow{2}{*}{} & \multicolumn{2}{c}{(a) 160 hours}  & \multicolumn{2}{c}{(b) 40 hours}   \\ 
                  & \textit{dev}   & \textit{test}  & \textit{dev}   & \textit{test}  \\ \hline
\textbf{SE}       & 34.50          & 34.51          & 17.41          & 15.44          \\ \hline
\textbf{ME}       & 35.35          & 35.49          & 23.30          & 20.40          \\ \hline
\textbf{CD}       & 33.06          & 33.65          & 23.53          & 20.87          \\ \hline
\textbf{CS}       & \textbf{35.84} & \textbf{36.32} & \textbf{23.54} & \textbf{21.72} \\ \bottomrule
\end{tabular}
\caption{BLEU scores trained on different size of data.}
\label{table:BLEU}
\end{table}

\noindent \textbf{High-resource}: Column (a) in Table~\ref{table:BLEU} showed the results trained on 160 hours of data. \textbf{CD} and \textbf{CS} represent the proposed methods mentioned in Sec.~\ref{sec:cosine} and~\ref{sec:cosine_softmax} respectively.
We got mixed results on further applying pre-trained word embedding on \textbf{ME}. 
\textbf{CD} degraded the performance, which is even worse than \textbf{SE}, but \textbf{CS} performed the best.
Results showed that directly learn word embedding via cosine distance is not a good strategy in the high-resource setting, but integrating similarity with cosine softmax function can significantly improve performance.
We leave the discussion in Sec.~\ref{subsubsec:discussion}.

\noindent \textbf{Low-resource}:
We also experimented on 40 hours subset data for training, as shown in column (b) in Table~\ref{table:BLEU}.
We could see that \textbf{ME}, \textbf{CD} and \textbf{CS} overwhelmed \textbf{SE} in low-resource setting.
Although \textbf{CD} resulted in degrading performance in high-resource setting, it showed improvements in low-resource scenario.
\textbf{CS} consistently outperformed \textbf{ME} and \textbf{CD} on different data size, showing it is robust on improving ST task.

\subsection{Analysis of Recognition Decoder Output}
In this section, we analyzed hidden states $\mathbf{s}$ by existing methods.
For each word $v$ in corpus, we denoted its word embedding $\hat{e}_{v}$ as \emph{pre-trained embedding}, and $e_v$ as \emph{predicted embedding}.
Note that because a single word $v$ could be mapped by multiple audio segments, we took the average of all its predicted embedding.
We obtained the top 500 frequent words in the whole Fisher Spanish corpus, and tested on the sentences containing only these words in \emph{test} set. 

\noindent \textbf{Eigenvector Similarity}: To verify our proposed methods can constrain hidden states in the word embedding space, we computed eigenvector similarity between \emph{predicted embedding} and \emph{pre-trained embedding} space.
The metric derives from Laplacian eigenvalues and represents how similar between two spaces, the \emph{lower} value on the metric, the \emph{more} approximately isospectral between the two spaces.
Previous works showed that the metric is correlated to the performance of translation task~\cite{sogaard2018limitations,chung2019towards}.
As shown in Table~\ref{table:eigen}, \emph{predicted embedding} is more similar to \emph{pre-trained embedding} when models trained on sufficient data (160 v.s 40 hours).
\textbf{CD} is the most similar case among the three cases, and \textbf{ME} is the most different case. 
Results indicated that our proposals constrain hidden states in \emph{pre-trained embedding} space. 

\begin{table}[t]
\centering
\begin{tabular}{c|cccc}
\toprule
\multirow{2}{*}{} & \multicolumn{2}{c}{160 hours} & \multicolumn{2}{c}{40 hours} \\
                  & \textit{dev}  & \textit{test}  & \textit{dev}  & \textit{test} \\ \hline
\textbf{ME}       & 16.50         & 18.58          & 13.80         & 15.09         \\ \hline
\textbf{CD}       & \textbf{2.60}          & \textbf{3.44}           & \textbf{3.95}          & \textbf{3.63}          \\ \hline
\textbf{CS}       & 11.55         & 13.76          & 8.62          & 9.80          \\
\bottomrule
\end{tabular}
\caption{Eigenvector similarity.}
\label{table:eigen}
\end{table}

\begin{table}[]
\centering
\begin{tabular}{c|cccc}
\toprule
\multirow{2}{*}{} & \multicolumn{2}{c}{160 hours}  & \multicolumn{2}{c}{40 hours}   \\ 
                  & P@1            & P@5            & P@1            & P@5            \\ \hline
\textbf{ME}       & 1.85           & 6.29           & 1.11           & 9.62           \\ \hline
\textbf{CD}       & \textbf{61.48} & \textbf{77.40} & \textbf{56.30} & \textbf{69.25} \\ \hline
\textbf{CS}       & 17.78          & 35.19          & 10.37          & 25.19          \\ \bottomrule
\end{tabular}
\caption{Precision@k of semantic alignment on \emph{test} set.}
\label{table:wordtranslation}
\end{table}

\noindent \textbf{Semantic Alignment}: To further verify if \emph{predicted embedding} is semantically aligned to \emph{pre-trained embedding}, we applied
Procrustes alignment~\cite{conneau2017word,lample2017unsupervised} method to learn the mapping between \emph{predicted embedding} and \emph{pre-trained embedding}.
Top 50 frequent words were selected to be the training dictionary, and we evaluated on the remaining 450 words with cross-domain similarity local scaling (CSLS) method. Precision@k (P@k, k=1,5) were reported as measurements.
As shown in Table~\ref{table:wordtranslation}, \textbf{CD} performed the best, and \textbf{ME} was the worst one.
This experiment reinforced that our proposals can constrain hidden states to the similar structure of word embedding space. 

\subsection{Speech Recognition Evaluation}

We further analyzed the results of speech recognition for \textbf{ME} and \textbf{CS}.
To obtain the recognition results from Eq~(\ref{eq:cos_softmax}), simply take $\argmax_v P_{CS}(v)$.
The word error rate (WER) of the source language recognition was reported in Table~\ref{table:wer}. 
Combining the results shown in Table~\ref{table:BLEU}, we could see that \textbf{CS} has worse WER, but higher BLEU compared with \textbf{ME}. 
We concluded that although leveraging word embedding at the intermediate level instead of text results in worse performance in speech recognition (this indicates that the WER of the recognition part does not fully determine the translation performance), the semantic information could somewhat help multitask models generate better translation in terms of BLEU. 
We do not include the WER of CD in Table~\ref{table:BLEU} because its WER is poor ($>$100\%), but interestingly, the BLEU of CD is still reasonable, which is another evidence that WER of the intermediate level is not the key of translation performance.

\subsection{Cosine Distance (CD) v.s. Softmax (CS)} 
\label{subsubsec:discussion}
Based on experimental results, we found that proposals are possible to map speech to semantic space.
With optimizing \textbf{CS}, BLEU consistently outperformed \textbf{ME}, which shows that utilizing semantic information truly helps on ST.
Directly minimizing cosine distance made the \textit{predicted embedding} space closest to \textit{pre-trained embedding} space, but performed inconsistently on BLEU in different data sizes.
We inferred that the imbalance word frequency training and hubness problem~\cite{faruqui2016problems} in word embedding space made hidden states not discriminated enough for the target language decoder while optimizing \textbf{CS} can alleviate this issue. 

\begin{table}[t]
\centering
\begin{tabular}{c|cccc}
\toprule
            & \multicolumn{2}{c}{160 hours} & \multicolumn{2}{c}{40 hours} \\
            & \textit{dev}  & \textit{test} & \textit{dev}  & \textit{test} \\ \hline
\textbf{ME} & 43.13          & 38.57          & 53.42          & 54.70              \\ \hline
\textbf{CS} & 50.15          & 44.43          & 57.63          & 57.21          \\ \bottomrule
\end{tabular}
\caption{Word error rate (\%) trained on different size of data.}
\label{table:wer}
\end{table}

\vspace{-1.5mm}
\section{Conclusions}
\label{sec:conclusion}
\vspace{-2mm}
Our proposals showed that utilizing word embedding as intermediate helps with the ST task, and it is possible to map speech to the semantic space.
We also observed that lower WER in source language recognition not imply higher BLEU in target language translation.

This work is the first attempt to utilize word embedding in the ST task, and further techniques can be applied upon this idea.
For example, cross-lingual word embedding mapping methods can be considered within the ST model to shorten the distance between MT and ST tasks.


\bibliographystyle{acl_natbib}
\bibliography{acl2020}

\appendix

\section{Appendix}
\label{sec:appendix}
\subsection{Single-task end-to-end model}
\label{subsec:SE_method}
One of our baseline models is a single-task end-to-end model, which is abbreviated as \textbf{SE} in the previous section. \textbf{SE} was trained using the source language speech and the target language text. It shares the same architecture with the multitask model but without the source language text decoding (without the recognition part in Fig.~\ref{fig:triangle_model}(a)).  
And its objective function can be written as:
\begin{equation} \label{eq:singletask_loss}
    \begin{split}
    \mathcal{L}_\text{SE} =  \mathcal{L}_\text{tgt} = \sum_{q} -\log P(y_q).\\
    \end{split}
\end{equation}
Further details can be referred to \cite{anastasopoulos-chiang-2018-tied}.

\subsection{Using different Word Embeddings}
\label{subsec:word_emb} 
Our proposed model benefits from publicly available pre-trained word embedding, which is easy-to-obtain yet probably coming from the domains different from testing data. It can bring to ST models in a simple plug-in manner.

In Sec.~\ref{subsec:evaluations}, we used word embedding trained on Wikipedia. To demonstrate the improvement of using different word embeddings, we additionally provide results of ST models using word embeddings trained on Fisher Spanish corpus (\textit{train} and \textit{dev} set) in Table~\ref{table:different_wemb_bleu}. Here we use the abbreviation of word embedding trained on Wikipedia as W-emb and word embedding trained on Fisher Spanish corpus as F-emb.

In \textbf{CD}/\textbf{CS} method, using F-emb obtained 0.27/0.61 improvement from using W-emb on \textit{dev} set.
And, \textbf{CD} got 0.15 improvement but \textbf{CS} got 0.51 degrading performance on \textit{test} set. 

The improvements show that using word embeddings trained in the related domain helps on the performance. 
In \textbf{CD} method, although using F-emb improves the performance, it still under-performed \textbf{ME} method.
It indicates that the selection of adopting methods is critical. 
In \textbf{CS} method, it got a great improvement on \textit{dev} set but not on \textit{test} set. 
It shows that using F-emb does help with the performance, but using word embedding trained on rich data (W-emb) could provide additional information that can generally extend to the \textit{test} set.

In general, whether using F-emb or W-emb as the training target, the experimental results show consistency to the discussion in Sec.~\ref{subsec:evaluations}.

\begin{table}[]
\centering
\begin{tabular}{@{}c|l|cc@{}}
\toprule
\multirow{2}{*}{} &
  \multicolumn{1}{c|}{\multirow{2}{*}{\begin{tabular}[c]{@{}c@{}}Word Embedding\\ Source\end{tabular}}} &
  \multicolumn{2}{c}{160 hours} \\ \cmidrule(l){3-4} 
                             & \multicolumn{1}{c|}{} & \textit{dev} & \textit{test} \\ \midrule
\multicolumn{1}{l|}{\textbf{ME}} &
  \multicolumn{1}{c|}{-} &
  \multicolumn{1}{l}{35.35} &
  \multicolumn{1}{l}{35.49} \\ \midrule
\multirow{2}{*}{\textbf{CD}} & Wikipedia             & 33.06        & 33.65         \\
                             & Fisher Spanish        & 33.33        & 33.80          \\ \midrule
\multirow{2}{*}{\textbf{CS}} & Wikipedia             & 35.84        & 36.32         \\
                             & Fisher Spanish        & 36.45        & 35.81         \\ \bottomrule
\end{tabular}
\caption{BLEU scores on using different pre-trained word embeddings.}
\label{table:different_wemb_bleu}
\end{table}

\end{document}